\documentclass{article}

\usepackage[preprint]{tmlr}
\setcitestyle{numbers,square,citesep={,}}

\usepackage{microtype}
\usepackage{graphicx}
\usepackage{caption}
\usepackage{subcaption}
\usepackage{wrapfig}
\usepackage{booktabs} \usepackage{scalefnt}

\usepackage{amsmath}
\usepackage{amssymb}
\usepackage{mathtools}
\usepackage{amsthm}
\usepackage{amsfonts}
\usepackage{bbold}
\usepackage{enumitem}
\usepackage{cancel}
\usepackage{graphicx}
\usepackage{comment}
\usepackage{relsize}
\usepackage{xcolor}
\usepackage{aliascnt}

\usepackage{hyperref}

\usepackage[capitalize,noabbrev]{cleveref}

\theoremstyle{plain}

\newaliascnt{Thm}{sa}
\newtheorem{Thm}[Thm]{Theorem}
\aliascntresetthe{Thm}

\newaliascnt{Lem}{sa}
\newtheorem{Lem}[Lem]{Lemma}
\aliascntresetthe{Lem}

\newaliascnt{Prp}{sa}
\newtheorem{Prp}[Prp]{Proposition}
\aliascntresetthe{Prp}

\newaliascnt{Prf}{sa}
\newtheorem{Prf}[Prf]{Proof}
\aliascntresetthe{Prf}

\newaliascnt{Cor}{sa}

\aliascntresetthe{Cor}

\newaliascnt{Con}{sa}

\aliascntresetthe{Con}

\newaliascnt{Fct}{sa}

\aliascntresetthe{Fct}

\newaliascnt{Prn}{sa}

\aliascntresetthe{Prn}

\newaliascnt{Def}{sa}
\newtheorem{Def}[Def]{Definition}
\aliascntresetthe{Def}

\newaliascnt{DefLem}{sa}

\aliascntresetthe{DefLem}

\newaliascnt{Axm}{sa}

\aliascntresetthe{Axm}

\newaliascnt{Not}{sa}
\newtheorem{Not}[Not]{Notations}
\aliascntresetthe{Not}

\newaliascnt{Asm}{sa}

\aliascntresetthe{Asm}

\newaliascnt{Rem}{sa}
\newtheorem{Rem}[Rem]{Remark}
\aliascntresetthe{Rem}

\newaliascnt{Cau}{sa}

\aliascntresetthe{Cau}

\newaliascnt{Eg}{sa}

\aliascntresetthe{Eg}

\newaliascnt{Tho}{sa}

\aliascntresetthe{Tho}

\newcommand{\N}{\mathbb{N}}
\newcommand{\R}{\mathbb{R}}

\newcommand{\Bcal}{\mathcal{B}}

\newcommand{\Fcal}{\mathcal{F}}

\newcommand{\Hcal}{\mathcal{H}}

\newcommand{\Pcal}{\mathcal{P}}

\newcommand{\Xcal}{\mathcal{X}}
\newcommand{\Ycal}{\mathcal{Y}}
\newcommand{\Zcal}{\mathcal{Z}}

\newcommand{\Ev}{\textsc{e}}

\newcommand{\Pbb}{\mathbb{P}}

\newcommand{\ins}{\subseteq}

\newcommand{\E}{\mathbb{E}}

\DeclareMathOperator{\KL}{KL}

\DeclareMathOperator*{\argmax}{argmax}

\DeclareMathOperator*{\argmin}{argmin}

\newcommand{\lp}{\left ( }
\newcommand{\rp}{\right ) }

\newcommand{\lB}{\left [ }
\newcommand{\rB}{\right ] }
\newcommand{\lC}{\left \{ }
\newcommand{\rC}{\right \} }
\newcommand{\lI}{\left| }
\newcommand{\rI}{\right| }

\newcommand{\st}{\,\middle|\,}

\usepackage{algorithm}
\usepackage[noend]{algpseudocode}
\algnewcommand{\Input}[1]{\State \textbf{Input:}
	\Statex \hspace*{\algorithmicindent}\parbox[t]{.8\linewidth}{\raggedright #1}
}
\algnewcommand{\Initialize}[1]{\State \textbf{Initialize:}
	\Statex \hspace*{\algorithmicindent}\parbox[t]{.8\linewidth}{\raggedright #1}
}
\algnewcommand{\Output}[1]{\State \textbf{Output:}
	\Statex \hspace*{\algorithmicindent}\parbox[t]{.8\linewidth}{\raggedright #1}
}\algrenewcommand\alglinenumber[1]{\tiny #1:}

\title{Anytime-valid simulation-based hypothesis testing}

\author{\name Patrick Forr\'e \email p.d.forre@uva.nl\\
        \addr  University of Amsterdam
        \AND
        \name Lydia Brenner \email lbrenner@nikhef.nl\\
        \addr  Nikhef
   }

\def\month{07}  \def\year{2026} \def\openreview{\url{https://openreview.net/forum?id=xxxxxxxx}}

\begin{document}

\maketitle

\begin{abstract}
For a given data distribution $(X_t)_{t \in \N} \overset{\text{i.i.d.}}{\sim} Q$, we investigate the hypothesis testing problem: $H_0: Q = P_0$ vs.\ $H_1: Q = P_1$, for two different model probability distributions $P_0$ and $P_1$. In contrast to the standard setting, where analytic densities $p_0$ and $p_1$ are given, here, we consider the density-free setting, where we only have access to simulations $(Z^0_t)_{t \in \N} \overset{\text{i.i.d.}}{\sim} P_0$ and $(Z^1_t)_{t \in \N} \overset{\text{i.i.d.}}{\sim} P_1$.
 For this simulation-based hypothesis testing setting, we construct an \Ev-test martingale, resulting in a sequential test with anytime-valid type-I error guarantees, approximate growth optimality, geometrically decaying type-II error bounds, and asymptotic power one. 
 Most ingredients used in our constructions are variants of well known concepts. The value of this paper lies in the compact presentation of an effective, anytime-valid solution for the density-free simulation-based sequential hypothesis testing case.
\end{abstract}

\section{Introduction}
\label{sec:intro}

Experimental particle physics generally focuses on distinguishing between competing hypotheses about the underlying processes responsible for the observed data. Often this is done in the form of testing an alternate hypothesis $H_1$ versus the so called \emph{Standard Model} (SM) of particle physics, which acts as the null hypothesis $H_0$. At the Large Hadron Collider (LHC), this is done both for searches for new phenomena and in precision measurements of the Standard Model. The observed events are the result of a complex sequence of stochastic processes, particle interactions with the detector, and the reconstruction of measurable quantities. Statistical hypothesis testing then provides the framework through which theoretical predictions are tested based on actual experimental observations  $x=(x_n)_{n \in [N]}$ that originated from the underlying true physical distribution $Q$. 

In particle physics, the alternative hypothesis $H_1$ usually consists only of a single model $P_1$ that arises as an extended version of the Standard Model $P_0$. 
In such a setting, where one only has a simple null hypothesis $H_0: Q=P_0$ and a simple alternative $H_1: Q=P_1$, the standard way of testing $H_0$ versus $H_1$ is through the likelihood ratio:
\begin{align}
    \Lambda(x) &:= \frac{p_1(x)}{p_0(x)},
\end{align}
where $p_1$ and $p_0$ are the corresponding probability densities of $H_1$ and $H_0$, respectively. The reason for this lies in the Neyman-Pearson lemma, which states that, for the simple-vs-simple setting, a test based on a threshold applied to the likelihood ratio $\Lambda(x)$ is the most powerful test at each fixed significance level $\alpha \in (0,1]$, see \cite{10.1098/rsta.1933.0009, Feldman:1997qc}. 

For example, in the LHC experiments such as ATLAS \cite{Aad:1129811} and CMS \cite{Chatrchyan:1129810}, in the search for the Higgs boson, which was discovered in 2012, see \cite{ATLAS2012Higgs, CMS2012Higgs}, a likelihood ratio was used between the null-hypothesis $H_0$ of ``no Higgs boson present'' and the alternate hypothesis $H_1$ of the ``existence of a Higgs boson'' in the data. 
In collider analyses, related likelihood-ratio and profile-likelihood constructions form the basis of widely used procedures for hypothesis testing, parameter estimation, and the treatment of nuisance parameters, see \cite{Cowan_2011,Prosper:1306523}.

However, in realistic LHC analyses, the probability densities $p_0$ and $p_0$ entering the likelihood ratio $\Lambda$ are rarely available in analytic or closed form. The mapping from the underlying theory parameters to the reconstructed detector-level data involves multiple stages of particle production and detector response, making direct evaluation of the high-dimensional likelihood ratio computationally difficult or intractable, see \cite{brehmer2020simulationbasedinferencemethodsparticle}. 
Monte Carlo simulation therefore plays a central role in modern collider physics. Event generators model the underlying particle interactions and subsequent evolution of the event, while detailed detector simulations model the propagation of particles through the experimental apparatus and its response. 
The resulting simulated events provide samples from the distributions predicted by a given physical hypothesis and allow theoretical models to be compared with data at the level at which measurements are actually performed. In consequence, simulations are used not only for estimating signal and background distributions but also for modelling detector efficiencies and acceptance, validating analysis procedures, and evaluating systematic uncertainties.
The connection between likelihood-ratio optimality and simulation is particularly relevant when the likelihood itself cannot be evaluated explicitly. If independent samples can be generated under competing hypotheses, the problem of distinguishing them can instead be formulated directly in terms of the simulated distributions.

With more modern methods, such as deep learning, a classifier can be trained to distinguish samples drawn from $P_0$ and $P_1$, under appropriate conditions and with sufficient capacity and training data, to learn an approximation $\hat \Lambda$ of the likelihood ratio $\Lambda$, which then can be used as a surrogate in the testing procedure and which allows to calculate p-values as a measure of significance.

However, the classical hypothesis testing approach based on p-values has further drawbacks, especially when the sample size needed for a targeted effect size is a priori unknown. The test is generally only valid for a chosen sample size fixed in advance: no ``peeking'', no optional stopping or optional continuation is allowed. In addition, one can not easily aggregate evidence against the null hypothesis $H_0$ from different experiments, or conduct multiple hypothesis tests at once, etc. 
As a remedy, the anytime-valid sequential testing procedure \emph{testing by betting} was recently developed, see \cite{Sha21, Vovk21, Gru20, Was20, Hen21, Ram23, Ram24book}. Also in this framework, the standard approaches require, in the simple-vs-simple setting, the likelihood ratio $\Lambda$, which we do not have in the density-free, simulation-based context.

In this paper, we want to address both points at once: 1.) the density-free/simulation-based hypothesis testing setting (without analytic densities $p_1$, $p_0$), and: 2.) building an anytime-valid sequential hypothesis testing procedure. Our construction will result in a sequential test, which comes with anytime-valid type-I error guarantees, approximate growth optimality, geometrically decaying type-II error bounds and asymptotic power one. We also show that the test statistics will implicitly approximate the likelihood ratio $\Lambda$.

\section{Anytime-valid hypothesis testing with conditional \Ev-variables}

In this section, we start by giving an overview over the main results from the standard literature of anytime-valid hypothesis testing, see, for instance, \cite{Sha21, Vovk21, Gru20, Was20, Hen21, Ram23, Ram24book}, etc.

In the following, let $\Omega$ be a measurable space with $\sigma$-algebra $\Bcal$, which we consider the underlying sample space, on which all occurring random variables live. We will endow $(\Omega,\Bcal)$ with several different probability distributions: $Q$, the true distribution, $P_0$ the standard model distribution, and $P_1$ the alternative model distribution. 
Based on a (finite or infinite) sequence of data points $(X_t)_{t \in \N} \sim Q$ we want to test the hypothesis if the true distribution $Q$ rather coincides with the standard model distribution $P_0$ or with the alternative model distribution $P_1$:
\begin{align}
    H_0: Q= P_0, \qquad \text{ vs. } \qquad  H_1: Q=P_1.
\end{align}
The safe hypothesis testing framework is based on the following concepts:

\begin{Def}[\Ev-variables]
    A non-negative measurable map $E:\Omega \to [0,\infty]$ is called an \emph{\Ev-variable for $H_0$} if:
    \begin{align}
        \E_{P_0}[E] \le 1.
    \end{align}
\end{Def}

Note that the Markov inequality shows that every \Ev-variable $E$ for $H_0$ induces a p-variable $\pi:=\min(1,1/E)$ for $H_0$:
\begin{align}
 P_0( \pi \le \alpha) = P_0 \lp E \ge \frac{1}{\alpha} \rp     \overset{!}{\le} \alpha, \qquad \forall \alpha \in (0,1].
\end{align}
In this sense, \Ev-variables can be considered more restrictive, so called \emph{post-hoc}, (inverses of) p-variables, see \cite{Kon23, Gru24, Ram24book, Chu26}, and, they directly produce a decision rule for every confidence level $\alpha \in (0,1]$: ``Reject $H_0$ if $E \ge \frac{1}{\alpha}$.''. The inequality above shows that the type-I error is then bounded by $\alpha$. We remain to investigate the type-II error, the \emph{power}, resp., for $H_1$. 
However, in contrast to the classical concept of \emph{the power}  $\beta:=P_1\lp \pi \le c \rp$ of a test, which depends on the chosen threshold $c$, in the safe hypothesis testing framework, the \emph{Kelly criterion} is used instead to introduce the \emph{\Ev-power}, which quantifies on a logarithmic scale how much evidence one can expect to gather against $H_0$ under the alternative model $P_1$.

\begin{Def}[\Ev-power]
    Let $E:\Omega \to [0,\infty]$ be an \Ev-variable for $H_0$. Then its \emph{\Ev-power} for $H_1$ is given by:
    \begin{align}
       \E_{P_1}[\log E].
    \end{align}
\end{Def}

In the simple-vs-simple hypothesis test situation, which we are considering in this paper, we can even point out the most \Ev-powerful \Ev-variable, which directly follows from the Donsker-Varadhan variational formula for the Kullback-Leibler divergence, see \cite{Don75I, Don76III, Dup97} or \cite{Sha21}:

\begin{Lem}[Neyman-Pearson lemma for the optimal \Ev-power]
\label{lem:neyman-pearson-e-power}
    Under the assumption of absolute continuity: $P_1 \ll P_0$,
     the $P_0$-essentially unique \Ev-variable for $H_0$ with maximal \Ev-power for $H_1$ is given by the likelihood-ratio, i.e.\ the Radon-Nikodym derivative:
    \begin{align}
        E^*(x) &:= \frac{dP_1}{dP_0}(x),
    \end{align}
    which has the \Ev-power:
    \begin{align}
          \E_{P_1}[\log E^*] \overset{!}{=} \KL(P_1\|P_0),
    \end{align}
    the Kullback-Leibler divergence between $P_1$ and $P_0$.
\end{Lem}

Since, in this paper, we are in the situation, where the ratio $\frac{dP_1}{dP_0}$ is not given to us in an analytic form or as an algorithmic function, we cannot directly use it in practice as our \Ev-variable. In the next section we thus aim to construct other \Ev-variables for $H_0$ with an \Ev-power close to the maximal one of $\KL(P_1\|P_0)$ and which are only based on simulations from $P_0$ and $P_1$.

Before we go there, first note, that the concepts of \Ev-variable $E$ and its \Ev-value $E(x)$, in the presented form, are only restricted to either one realization $x \sim Q$ or an array of data points of fixed given length $x=(x_1,\dots,x_t)$ with $x_i \sim Q$. To turn this into a flexible framework where the number of data points $x_t \sim Q$ can grow over time $t \to \infty$ and where we can build on information from the past, we have to introduce the following concept:

\begin{Def}[Conditional \Ev-variables and conditional \Ev-power]
    Let $\Fcal \ins \Bcal$ be a sub-$\sigma$-algebra of $\Bcal$\footnote{A sub-$\sigma$-algebra $\Fcal \ins \Bcal$ is typically of the form $\Fcal=\sigma(Z_1,\dots,Z_t)$ for some random variables $Z_1,\dots,Z_t: \Omega \to \Zcal$. So, $\Fcal$ then encodes the information of the outcomes of $Z_1,\dots,Z_t$.} on $\Omega$. 
    A non-negative measurable map $E:\Omega \to [0,\infty]$ is called a \emph{conditional \Ev-variable for $H_0$ conditional on $\Fcal$} if:
    \begin{align}
        \E_{P_0}[E|\Fcal] \le 1 \qquad P_0\text{-a.s.}
    \end{align}
    Its \emph{conditional \Ev-power} for $H_1$ conditional on $\Fcal$ is defined by:
    \begin{align}
        \E_{P_1}[\log E|\Fcal].
    \end{align}
\end{Def}

Note that a conditional \Ev-variable for $H_0$ conditional to $\Fcal$ is also a (unconditional) \Ev-variable for $H_0$. More generally, we can consider a sequence of $(E_t)_{t \in \N_0}$ conditional \Ev-variables for $H_0$, each constructed by using past information. Formally, we introduce:

\begin{Not}
\label{not:main}
   Let $(E_t)_{t \in \N_0}$ be a sequence of conditional \Ev-variables for $H_0$, $E_t: \Omega \to [0,\infty]$, $E_0:=1$, adapted to the filtration $\Fcal=(\Fcal_t)_{t \in \N_0}$ of sub-$\sigma$-algebras $\Fcal_t \ins \Bcal$:
   \begin{align}
       \Fcal_0 \ins \Fcal_1 \ins \Fcal_2 \ins \cdots \ins \Fcal_t \ins \cdots \ins \Bcal,
   \end{align}
   each conditional on its own past, that is, for every $t \in \N_1$ the map $E_t: \Omega \to [0,\infty]$ is $\Fcal_t$-measurable and:
    \begin{align}
        \E_{P_0}[E_t|\Fcal_{t-1}] \le 1 \qquad P_0\text{-a.s.}
    \end{align}
    We further construct non-negative variables $W_t:\Omega \to [0,\infty]$ as follows for $t \in \N_1$ by multiplication:
    \begin{align}
         W_t &:= E_t \cdot W_{t-1} = E_t \cdot E_{t-1}\cdots E_1, & W_0 &:= 1.
    \end{align}
\end{Not}

With these notations we have the following:

\begin{Lem}
    With the \Cref{not:main}, each $W_t$ for $t \ge 0$ is an (unconditional) \Ev-variable for $H_0$ and the sequence $(W_t)_{t \in \N_0}$ is a non-negative supermartingale adapted to the filtration $\Fcal=(\Fcal_t)_{t \in \N_0}$ w.r.t.\ $P_0$. More concretely, for $t \in \N_1$ we have:
    \begin{align}
        \E_{P_0}[W_t|\Fcal_{t-1}] &\le W_{t-1} \qquad P_0\text{-a.s.}, &  \E_{P_0}[W_0]&=1.
    \end{align}
\end{Lem}

The above sequence $(W_t)_{t \in \N_0}$ now allows us to define a \emph{sequential decision rule} with $\alpha \in (0,1]$ as follows:
\begin{center}
   ``Reject $H_0$ if there \emph{ever} appears any $t \in \N_0$ with $W_t > \frac{1}{\alpha}$.''
\end{center}
Remarkably, this sequential decision rule has the following \emph{anytime-valid type-I error guarantee},  provided by Ville's inequality for non-negative supermartingales, see \cite{Vil39} or \Cref{thm:ville}:

\begin{Thm}[Anytime-valid type-I error bounds]
\label{thm:type-I-error}
    With the \Cref{not:main}, we have the following upper bound on the \emph{anytime-valid type-I error} for the above sequential test for every $\alpha \in (0,1]$:
    \begin{align}
        P_0\lp \exists t \in \N_0.\; W_t > \frac{1}{\alpha} \rp = P_0\lp \sup_{t \in \N_0} W_t > \frac{1}{\alpha} \rp  \overset{!}{\le} \alpha.
    \end{align}
\end{Thm}

Under boundedness conditions on the logarithms of the $E_t$'s and sufficiently big conditional \Ev-power, we also get an \emph{asymptotic type-II error control} with help of the Azuma–Hoeffding inequality for bounded,  filtration adapted sequences of random variables, see \cite{Azu67,For26b} or \Cref{thm:azuma}:

\begin{Thm}[Geometrically decaying type-II error bounds for log-bounded conditional \Ev-variables]
\label{thm:type-II-error}
With the \Cref{not:main}, assume that for every $t \in \N_1$ there are constants $a_t, b_t, c_t \in \R$ such that:
\begin{align}
    a_t &\le \log E_t \le b_t \quad P_1\text{-a.s.}, & 
    c_t & \le \E_{P_1}[\log E_t|\Fcal_{t-1}]
    \quad P_1\text{-a.s.}.
\end{align}
Then for every $\alpha \in (0,1]$ and $t \in \N_1$ we have the following type-II error bound\footnote{For $s \in \R$ we put $s_+:=\max(0,s)$ and $s_+^2:=(s_+)^2$.}:
\begin{align}
P_1\lp W_t\le \frac{1}{\alpha} \rp & 
    \le \exp\lp - \frac{2\lp \sum_{n=1}^t c_n - \log \frac{1}{\alpha}\rp_+^2}{\sum_{n=1}^t (b_n-a_n)^2} \rp.
\end{align}
In particular, if the bounds are the same ($a_t=a$, $b_t=b$, $c_t=c$) for all $t \in \N_1$ with positive (conditional) \Ev-power lower bound $c >0$ for $H_1$ then the \emph{type-II error} bound simplifies and converges to zero with asymptotically \emph{geometric rate}:
\begin{align}
    P_1\lp W_t\le \frac{1}{\alpha} \rp & 
    \le \exp\lp - 2t \cdot \frac{\lp c - \frac{1}{t}\log \frac{1}{\alpha}\rp_+^2}{(b-a)^2} \rp \overset{t \to \infty}{\longrightarrow} 0.
\end{align} 
As a consequence, we then have \emph{asymptotic power one}:
\begin{align}
   P_1\lp \sup_{t \in \N_0} W_t > \frac{1}{\alpha} \rp = P_1\lp \sup_{t \in \N_0} W_t = \infty \rp & = 1.
\end{align}
\end{Thm}

\begin{Rem}[Clipping log-unbounded \Ev-variables]
\label{rem:clipping}
    If $E$ is an \Ev-variable for $H_0$, but for which $\log E$ might not be bounded, then for every $\gamma \in (0,1]$ we can introduce its \emph{$\gamma$-clipped} version:
    \begin{align}
        E_\gamma(x) &:= \gamma + (1-\gamma) \cdot \min(E(x),\gamma^{-1}) \le 1 + E(x),
    \end{align}
    which is again an \Ev-variable for $H_0$ and for which we then have the bounds: 
    \begin{align}
        - \log \frac{1}{\gamma} \le \log E_\gamma(x) \le \log \frac{1}{\gamma},
    \end{align}
    together with the pointwise convergence:
    \begin{align}
        \lim_{\gamma \to 0} E_\gamma(x) &= E(x),
    \end{align}
    which is monotone non-decreasing if $E(x)\ge 1$ and monotone non-increasing if $E(x) \le 1$ for $\gamma \to 0$.
    This then implies the following convergences of the integrals for arbitrary measures $P$:
    \begin{align}
        \lim_{\gamma \to 0} \int E_\gamma\, dP &= \int E\, dP, & \lim_{\gamma \to 0} \int \log E_\gamma\, dP &= \int \log E\, dP.
    \end{align}
    The latter limit holds, whenever the integral on the right is well-defined (possibly $\pm \infty$, but excluding the $\infty - \infty$ case). The convergence of the integrals can be shown with help of the monotone convergence theorem, applied on the sets $\lC E \le 1 \rC$ and $\lC E > 1 \rC$ separately.

    This allows us to take an \Ev-variable $E$ for $H_0$ with positive \Ev-power $\E_{P_1}[\log E] > 0$ for $H_1$ (but with possible unbounded $\log E$), and clip it with some small $\gamma > 0$. If $\gamma >0$ is sufficiently small then also $E_\gamma$ will have positive \Ev-power $\E_{P_1}[\log E_\gamma] >0$ for $H_1$ and, in addition, the geometrically decaying type-II error bounds of \Cref{thm:type-II-error} will hold with the bounds $a:= - \log \frac{1}{\gamma}$, $b:=\log \frac{1}{\gamma}$ and $c:=\E_{P_1}[\log E_\gamma] >0$.
\end{Rem}

\begin{Rem}[Rejection time]
\label{rem:rejection-time}
    With the \Cref{not:main}, we introduce the \emph{rejection time} of the test supermartingale $(W_t)_{t \in \N}$ w.r.t.\ $\alpha \in (0,1]$:
    \begin{align}
        \tau_\alpha &:= \inf \lC t \in \N_0 \st W_t > \frac{1}{\alpha} \rC.
    \end{align}
    To get an intuition about the average rejection time, assume, for simplicity, that $(x_t)_{t \in \N} \sim P_1$ is i.i.d.\ and that we use the same \Ev-variable at each time: $E_t = E(x_t)$ for $t \in \N$. 
    Via Wald's equation, under integrability conditions\footnote{The minimal assumptions for the validity of Wald's equation are: $\E_{P_1}[(\log E)_-] < \infty$ and $\E_{P_1}[\log E] > 0$.}, see \cite{Wald44, Dur19}, we then get the following close approximation for the \emph{average rejection time}:
    \begin{align}
        \E_{P_1}[\tau_\alpha] = \frac{\E_{P_1}\lB \log W_{\tau_\alpha}\rB}{\E_{P_1}[\log E]} &  \;\gtrapprox\; \frac{\log \frac{1}{\alpha}}{\E_{P_1}[\log E]},
    \end{align}
    which can be interpreted as the average number of samples needed to obtain a rejection. 
    The above formula shows that, if we want to minimize the average rejection time, we need to maximize the \Ev-power of $E$, which by \Cref{lem:neyman-pearson-e-power} is uniquely maximized by the likelihood ratio: $E^*=\frac{dP_1}{dP_0}$, for which we then get the minimal average rejection time:
       \begin{align}
        \E_{P_1}[\tau^*_\alpha] & \;\gtrapprox\; \frac{\log \frac{1}{\alpha}}{\KL(P_1\|P_0)}.
    \end{align} 
\end{Rem}

\section{Construction of \Ev-variables in the density-free, simulation-based setting}

\label{sec:e-val-constr}

In this section, we now want to construct \Ev-powerful \Ev-variables $E$ for a sequential i.i.d.\ hypothesis testing problem:
\begin{align}
    H_0: Q= P_0, \qquad \text{ vs. } \qquad  H_1: Q=P_1,
\end{align}
where we do not have access to the densities of $P_0$ or $P_1$, or even just the likelihood ratio $\frac{dP_1}{dP_0}$. 
Instead, we aim to construct \Ev-variables that are constructed only from i.i.d.\ samples $(z^0_t)_{t \in \N} \sim P_0$ and $(z^1_t)_{t \in \N} \sim P_1$, but are close to the optimal \Ev-variable: $E^*:=\frac{dP_1}{dP_0}$.

To construct \Ev-variables, consider an arbitrary function: $h: \Omega \to [0,\infty]$ with $0< \E_{P_0}[h] < \infty$. Then a simple normalization turns $h$ into a valid \Ev-variable for $H_0$:
\begin{align}
    \tilde h(z) &:= \frac{h(z)}{\E_{P_0}[h]}, & \E_{P_0}[\tilde h] &=1. \label{eq:E-val-h-norm}
\end{align}
The problem is that we usually cannot analytically solve the integral: $\E_{P_0}[h] = \int h \, dP_0$. 
If we instead approximate this integral with i.i.d.\ samples from $z^0_1,\dots, z^0_K \sim P_0$, we get:
\begin{align}
\hat h_K(z) &:= \frac{h(z)}{\frac{1}{K}\sum_{k=1}^K h(z^0_k)}  \quad \underset{\text{a.s.}}{\overset{K \to \infty}{\longrightarrow}}  \quad \frac{h(z)}{\E_{P_0}[h]} = \tilde h(z).
\end{align}
However, this might not be a valid \Ev-variable for $H_0$ anymore for fixed finite $K \ge 1$, as expectation and quotients don't commute in general:
\begin{align}
 \E_{}\lB\hat h_K\rB = \E_{}\lB \frac{h(Z)}{\frac{1}{K}\sum_{k=1}^K h(Z^0_k)} \rB \overset{\text{i.g.}}{\neq} \frac{\E_{}\lB h(Z)\rB}{\E_{}\lB\frac{1}{K}\sum_{k=1}^K h(Z^0_k) \rB} = 1. 
\end{align}
A commonly known trick in statistics and machine learning is to turn such a quantity into a classification probability (or variants thereof), see
\cite{Qin98, Bic07, Sug08, Sug09, Ngu10, Kan10, Gut10,  Sug12, Good14, Cra15, Cra16, Lhe18, Bre18, Oord18, Hje19, Poo019, Her20, Dur20, Chen20, Was20, Kim21,  Fed20, Fed21, Fed23, Fed24, Mil21, Mil22, Mil23, Del22, Del23, Pan24a, Pan24b, Lip26, Domb26, Cor26, Bor26}\footnote{The papers \cite{Cor26, Bor26} use a similar construction and learnable approach to ours, with a somewhat different focus. \cite{Cor26} constructs confidence sequences, while \cite{Bor26} considers a multi-arm bandit setting.}, where it is known that a maximizer of the underlying classification task is proportional to the (log) density ratio $\frac{dP_1}{dP_0}$. More concretely, we simply add the nominator $h(z)$ to (the average of) the denominator above and introduce the quantity:
\begin{align}
    E^{(K,h)}(z^1,z^0_1,\dots, z^0_K) &:= \frac{h(z^1)}{\frac{1}{K+1}\lp h(z^1)+\sum_{k=1}^K h(z^0_k)\rp}. \label{eq:E-val-K-h}
\end{align}
If $H_0: Q=P_0$ is true, then $z^1 \sim P_0$ together with $z^0_1,\dots, z^0_K \sim P_0$ i.i.d.\ of the same distribution $P_0$. This joint i.i.d.\ assumption then implies the same distributions for each of the ratios:
\begin{align}
    \frac{h(z^1)}{h(z^1)+h(z^0_1) + \cdots + h(z^0_k)} 
    \sim \frac{h(z^0_1)}{h(z^1)+h(z^0_1) + \cdots + h(z^0_k)} \sim \cdots \sim \frac{h(z^0_K)}{h(z^1)+h(z^0_1) + \cdots + h(z^0_k)}.
\end{align}
In particular, each ratio has the \emph{same expectation value}:
\begin{align}
    \E_{}\lB \frac{h(Z^1)}{h(Z^1)+h(Z^0_1) + \cdots + h(Z^0_k)} \rB 
=\cdots = \E_{}\lB \frac{h(Z^0_K)}{h(Z^1)+h(Z^0_1) + \cdots + h(Z^0_k)}\rB,
\end{align}
while they \emph{all add up to one}:
\begin{align}
    \E_{}\lB \frac{h(Z^1)}{h(Z^1)+h(Z^0_1) + \cdots + h(Z^0_k)} \rB +  \cdots + \E_{}\lB \frac{h(Z^0_K)}{h(Z^1)+h(Z^0_1) + \cdots + h(Z^0_k)}\rB = 1,
\end{align}
which thus implies:
\begin{align}
    \E_{}\lB \frac{h(Z^1)}{h(Z^1)+h(Z^0_1) + \cdots + h(Z^0_k)} \rB = \frac{1}{K+1}.
\end{align}
Multiplying by $(K+1)$ then shows that $E^{(K,h)}$ is a valid \Ev-variable for $H_0$. 
Formally, note, that we implicitly endowed all probability distributions with the $K$-fold independent product of $P_0$ in the expectation value. So, we actually investigated the equivalent hypothesis testing problem:
\begin{align}
    H_0: Q \otimes P_0^{\otimes K} = P_0 \otimes P_0^{\otimes K} =: \Pbb_0, \qquad \text{ vs. } \qquad  H_1: Q\otimes P_0^{\otimes K}=P_1\otimes P_0^{\otimes K}=:\Pbb_1.
\end{align}
In the rest of this section, we will mostly stay in this setting, but we might move between these different views without further indication. 

To summarize the above reasoning, we formally state the following:
\begin{Lem}[\Ev-value construction]
    Let $K\ge 1$ and $h:\Omega \to [0,\infty)$ be any map for which $0 < h(Z^1) + \sum_{k=1}^{K} h(Z^0_k) < \infty$ $\Pbb_1$-almost-surely, then the map $E^{(K,h)}$ given by:
\begin{align}
    E^{(K,h)}(z^1,z^0_1,\dots, z^0_K) &:= \frac{h(z^1)}{\frac{1}{K+1}\lp h(z^1)+\sum_{k=1}^K h(z^0_k)\rp},
\end{align}    
    is a valid \Ev-variable for $H_0$ with:
\begin{align}
   E^{(K,h)} &\ge 0, & \E_{\Pbb_0}\lB E^{(K,h)} \rB &= 1.
\end{align}
\end{Lem}
We now want to investigate, which choice of $h$ actually does optimize \Ev-power for $H_1$. For this we have the following theoretical result:

\begin{Prp}[Optimal \Ev-power for fixed $K \ge 1$, see \Cref{prf:prp:optimal-e-power-k}]
\label{prp:optimal-e-power-k}
    Let $P_0$, $P_1$ be two probability distributions on $\Omega$ with $P_1 \ll P_0$ and $K \ge 1$ be a fixed natural number. 
    Then consider all measurable functions $h: \Omega \to [0,\infty)$ for which $0 < h(Z^1) + \sum_{k=1}^{K} h(Z^0_k) < \infty$ $\Pbb_1$-almost-surely, and, the following \Ev-power objective:
    \begin{align}
        J_K(h) &:= \E_{\Pbb_1}\lB \log E^{(K,h)} \rB.
    \end{align}
    Then each maximizer $h^*$ of $J_K$ is $P_0$-almost-surely of the form:
    \begin{align}
         h^*(z) &= c \cdot \frac{dP_1}{dP_0}(z),
    \end{align}
    for some constant $c \in \R_{>0}$. Conversely, those functions maximize $J_K$ (for all $K \ge 1$ simultaneously).  The optimal \Ev-variable of our given type is thus given by:
    \begin{align}
        E^{(K,*)}(z^1,z^0_1,\dots,z^0_K) 
        &:= \frac{ \frac{dP_1}{dP_0}(z^1)}{\frac{1}{K+1}\lp \frac{dP_1}{dP_0}(z^1) + \sum_{k=1}^K \frac{dP_1}{dP_0}(z^0_k)\rp} 
        \quad \underset{\text{a.s.}}{\overset{K \to \infty}{\longrightarrow}} \quad \frac{dP_1}{dP_0}(z^1) =: E^*(z^1),
    \end{align}
which (for each fixed $z^1$) converges $P_0^{\otimes \N_1}$-almost-surely to the overall most \Ev-powerful \Ev-variable $E^*(z^1)=\frac{dP_1}{dP_0}(z^1)$ for $H_0$ vs.\ $H_1$, see \Cref{lem:neyman-pearson-e-power}, for $K \to \infty$. 
Furthermore, for the \Ev-power of $E^{(K,*)}$ we have the following Jensen inequality:
\begin{align}
    \KL(P_1\|P_0)  \quad \ge \quad  \E_{\Pbb_1}[\log E^{(K,*)}] \quad\overset{!}{\ge}\quad \KL(P_1\|P_0) - \log\lp 1 + \frac{\chi^2(P_1\|P_0)}{K+1} \rp \quad \overset{K \to \infty}{\longrightarrow} \quad\KL(P_1\|P_0),
\end{align}
provided $\chi^2(P_1\|P_0) < \infty$.
If, even more, $\frac{dP_1}{dP_0}$ is bounded, then we have the \emph{uniform} (in $z^1$) $P_0^{\otimes \N_1}$-almost-sure convergence:
\begin{align}
   \sup_{z^1 \in \Omega} \lI \log E^{(K,*)}(z^1,z^0_1,\dots,z^0_K) - \log E^*(z^1) \rI \quad \underset{\text{a.s.}}{\overset{K \to \infty}{\longrightarrow}} \quad 0.
\end{align}
\end{Prp}

We now focus on a way to learn to approximate such optimal \Ev-variables with simulations from $P_1$ and $P_0$.

\begin{Not}
\label{not:empirical-e-val}
To approximate the most \Ev-powerful \Ev-variable for $H_0$ vs.\ $H_1$, we make use of a flexible, optimizable paramterized function class $g_\theta: \Omega \to \R$, for example, deep neural networks of some architecture, with parameters $\theta \in \Theta$. 
We then put $h_\theta(z) := \exp(g_\theta(z))$ to use inside $E^{(K,h)}$ and get the parameterized \Ev-variables for $H_0$:
\begin{align}
    E^{(K,\theta)}(z^1,z^0_1,\dots,z^0_K) &:= \frac{ \exp(g_\theta(z^1))}{\frac{1}{K+1}\lp \exp(g_\theta(z^1)) + \sum_{k=1}^K \exp(g_\theta(z^0_k)) \rp }. \label{eq:e-var-ratio}
\end{align}

We will optimize the \Ev-power of $E^{(K,\theta)}$ by using an empirical estimate of the \Ev-power objective $J_K$ from above, using i.i.d.\ samples from $P_0$ and $P_1$.
For this first sample sufficiently many:
\begin{align}
    (z^1_n)_{n \in [N]} &\overset{\text{i.i.d.}}{\sim} P_1, & (z^0_{n,k})_{\substack{n \in [N]\\ k \in [K]}} &\overset{\text{i.i.d.}}{\sim} P_0.
\end{align}
We then consider the empirical objective:
\begin{align}
    J_{K,N}(\theta) &:= \frac{1}{N}\sum_{n=1}^N \log E^{(K,\theta)}\lp z^1_n,(z^0_{n,k})_{k \in [K]} \rp.
\end{align}
We then maximize this objective to receive the following estimator:
\begin{align}
     \hat \theta_{K,N} &: \in \argmax_{\theta \in \Theta}  J_{K,N}(\theta),
\end{align}
or, more generally, a sequence of \emph{near-maximizing estimators} $(\hat \theta_{K,N})_{N \in \N_1}$ that might come with some \emph{optimization gap} $(\hat \xi_{K,N})_{N \in \N_1}$, e.g.\ a gradient optimizer stuck in a local, non-global, maximum, etc., satisfying:
\begin{align}
     J_{K,N}(\hat \theta_{K,N}) &\ge  \sup_{\theta \in \Theta}  J_{K,N}(\theta) - \xi_{K,N}, & \xi_K &:= \limsup_{N \to \infty} \xi_{K,N} \in \R_{\ge 0}.
\end{align}
The resulting \Ev-variable for $H_0$ is then:
\begin{align}
    E^{(K,{\hat \theta_{K,N}})}(z^1,z^0_1,\dots,z^0_K) &= \frac{ \exp(g_{\hat \theta_{K,N}}(z^1))}{\frac{1}{K+1}\lp \exp(g_{\hat \theta_{K,N}}(z^1)) + \sum_{k=1}^K \exp(g_{\hat \theta_{K,N}}(z^0_k)) \rp }.
\end{align}    
\end{Not}

We can use M-estimation theory under standard Wald-type conditions, see \cite{Wald1949, Vaa98}, to get the following results on the asymptotic approximately optimal \Ev-power of the \Ev-variable $E^{(K,{\hat \theta_{K,N}})}$ for $H_0$ vs.\ $H_1$ for the limit $N \to \infty$.

\begin{Thm}[Approximate growth optimality, see \Cref{prf:thm:approx-gro}]
\label{thm:approx-gro}
Let the setting be like in \Cref{not:empirical-e-val}. 
Assume $P_1 \ll P_0$ and let $g_\theta :\Omega \to \R$ be a family of $\Bcal$-measurable functions, $\theta \in \Theta$, with the following properties:
\begin{enumerate}[series=approx-gro-list]
    \item Compactness: $\Theta$ is a compact metric space. \label{it:compact}
    \item Continuity: for every $z \in \Omega$ the map $\theta \mapsto g_\theta(z)$ is continuous. \label{it:cont}
    \item Non-degeneracy: there exists a $\theta \in \Theta$ with $J_K(\theta) > -\infty$. \label{it:non-deg}
    \item Uniform integrable envelopes: \label{it:unif-int-envs}
    \begin{align}
        \E_{P_0}\lB \sup_{\theta \in \Theta} g_\theta(Z^0)_+ \rB & < \infty, &
        \E_{P_1}\lB \sup_{\theta \in \Theta} g_\theta(Z^1)_- \rB & < \infty.
    \end{align}
\end{enumerate}
Then, for fixed $K \ge 1$, we have the following \emph{uniform strong law of large numbers}  for the objective, that is the $\Pbb_1$-almost-sure convergence:
\begin{align}
 \sup_{\theta \in \Theta} \lI J_{K,N}(\theta) - J_K(\theta) \rI &\overset{N \to \infty}{\longrightarrow} 0 \qquad \Pbb_1\text{-a.s.}
\end{align}
Furthermore, for every near-maximizing sequence $(\hat \theta_{K,N})_{N \in \N_1}$ of $(J_{K,N})_{N \in \N_1}$ with optimization gaps $(\xi_{K,N})_{N \in \N_1}$ we have the $\Pbb_1$-almost-sure convergences\footnote{Note that the expectation value $\E_{\Pbb_1}$ integrates over the variables $(z^1,z^0_1,\dots,z^0_K)$ of $E^{(K,\hat \theta_{K,N})}$, but leaves the randomness from the fitting procedure in $\hat \theta_{K,N}$ intact. The ``$\Pbb_1$-a.s.'' statement is then about the randomness in $\hat \theta_{K,N}$.}:
\begin{align}
\liminf_{N \to \infty}\E_{\Pbb_1}[\log E^{(K,\hat \theta_{K,N})}] 
 &\ge \sup_{\theta \in \Theta} J_K(\theta) -\xi_K \qquad \Pbb_1\text{-a.s.}
\end{align}
Now consider the maximizers $g^*_c: \Omega \to \R \cup \lC - \infty \rC$ of $J_K$ from \Cref{prp:optimal-e-power-k} (in log space) for $c \in \R$:
\begin{align}
    g^*_c(z) &:= g^*(z) + c, & g^*(z) &:= \log \frac{dP_1}{dP_0}(z).
\end{align}
We now assume, in addition to the above, that we have the following properties:
\begin{enumerate}[resume=approx-gro-list]
    \item Approximating class: there exist $\theta \in \Theta$ and $\epsilon, c \in \R$ with $0 \le \epsilon < \infty$ such that: \label{it:approx-class}
    \begin{align}
        \sup_{z \in \Omega} \lI g_\theta(z) - g^*_c(z) \rI \le \epsilon.
    \end{align}
    \item Finite $\chi^2$-divergence: we have: $P_1 \ll P_0$ with $\chi^2(P_1\|P_0) < \infty$. \label{it:chi-sq-ft}
\end{enumerate}
Then we have \emph{approximate growth optimality} (AGRO) expressed as the following $\Pbb_1$-almost-sure inequalities:
\begin{align}
     \underbrace{\KL(P_1\|P_0)}_{\text{maximal \Ev-power}}  & \ge \underbrace{\liminf_{N \to \infty}   \E_{\Pbb_1}[\log E^{(K,\hat \theta_{K,N})}]}_{\text{asymptotic learned \Ev-power}} \\
     &\ge \underbrace{\KL(P_1\|P_0)}_{\text{maximal \Ev-power}} \underbrace{-\log\lp 1 + \frac{\chi^2(P_1\|P_0)}{K+1} \rp}_{\text{penalty for finite }K} \underbrace{-\frac{1}{2}\epsilon^2}_{\text{approximation gap}} \underbrace{- \xi_K}_{\text{optimization gap}}.
\end{align}
In particular, if $\KL(P_1\|P_0) > \frac{1}{2}\epsilon^2 + \sup_{K\in \N_1}\xi_K$, then we can find $K,N  \ge 1$ such that we achieve positive \Ev-power $E_{\Pbb_1}[\log E^{(K,\hat \theta_{K,N})}] >0$ for $H_1$ with high probability.
\end{Thm}

After training, we have thus found the \Ev-variable $E^{(K,\hat \theta_{K,N})}$ for $H_0$ with optimized \Ev-power for $H_1$ or at least with positive \Ev-power $E_{\Pbb_1}[\log E^{(K,\hat \theta_{K,N})}] >0$ for $H_1$ with high probability.

For \emph{newly} sampled points at each time step $t \ge 1$:
\begin{align}
    x_t & \sim Q,&
    (z^0_{t,k})_{k \in [K]} & \overset{\text{i.i.d.}}{\sim} P_0,
\end{align}
we can thus finally define the (conditional) $\Ev$-variable at time $t \ge 1$:
\begin{align}
    E_t &:= E^{(K,\hat \theta_{K,N})}(x_t,z^0_{t,1},\dots,z^0_{t,K}),
\end{align}
and recursively the e-test martingale for $t \ge 1$:
\begin{align}
  W_t&:= E_t \cdot W_{t-1}, &  W_0 &:= 1.
\end{align}
We thus reject $H_0$ in favor of $H_1$ whenever $W_t$ crosses the threshold $\frac{1}{\alpha}$ and we can stop. Recall, that in real physical experiments, the standard is to use the so called \emph{5$\sigma$-rule} with 
$\frac{1}{\alpha} \approx 3.5 \cdot 10^6$, that is: $\log \frac{1}{\alpha} \approx 15$.

\begin{Rem}[Deep neural networks, see \cite{Good16, Bis23, Mur22}]
    For deep neural networks $g_\theta$ the map:
    \begin{align}
        (\theta,z) & \mapsto g_\theta(z),
    \end{align}
    is continuous and one usually even has a Lipschitz type inequality, based on the used architecture:
    \begin{align}
        |g_\theta(z)| & \le C(\theta) \cdot (1+\|z\|),
    \end{align}
    with a constant $C(\theta)$ that is continuous in $\theta$.
    Furthermore, the class of deep neural networks is a universal approximator. So, on compact $\Omega$, we can adjust the architecture and the parameterization such that: 
    \begin{align}
        g_\theta & \approx g^*_c,
    \end{align}
    for some $\theta$ from some sufficiently big, but compact $\Theta$.
    So, the remaining envelope properties from \Cref{thm:approx-gro} reduce for deep neural networks to simple moment conditions:
    \begin{align}
        \E_{P_1}[\|Z^1\|] &< \infty, & \E_{P_0}[\|Z^0\|] &< \infty.
    \end{align}
So, for deep neural networks, in practice, we can argue that \Cref{thm:approx-gro} should hold on compact $\Omega$. For non-compact, but $\sigma$-compact, $\Omega$ one can exhaust the space $\Omega$ with compacta and apply \Cref{thm:approx-gro} individually on each compact subspace. We do not provide further technical details for this generalization here.
\end{Rem}

\begin{Rem}[Removing the ambiguity of a constant]
 In practice, we can remove the ambiguity of the constant $c \in \R$ in the optimal function in \Cref{prp:optimal-e-power-k} and \Cref{thm:approx-gro} by introducing the following regularizer:
\begin{align}
    R_{K,N}(\theta) &:= \lI \log \frac{1}{NK} \sum_{n=1}^N \sum_{k =1 }^K \exp(g_\theta(z^0_{n,k})) \rI^2,
\end{align}
which enforces to keep $c \approx 1$ (or, $c \approx 0$ in log-space, resp.). 
We then minimize the following joint objective with a small regularization parameter $\kappa >0$:
\begin{align}
\hat \theta_{K,N}:\in  \argmax_{\theta \in \Theta} \lp J_{K,N}(\theta) - \kappa \cdot R_{K,N}(\theta) \rp.
\end{align}
Alternatively, we can sample separate i.i.d.\ samples $(z^0_m)_{m \in [M]} \sim P_0$ with $M \gg 1$ \emph{after} training (with or without such regularization) and compute:
\begin{align}
    \hat c &:= \log \frac{1}{M}\sum_{m=1}^M \exp\lp g_{\hat \theta_{K,N}}(z^0_m) \rp,  & \tilde g_{\hat \theta_{K,N}}(x) &:= g_{\hat \theta_{K,N}}(x) - \hat c.
\end{align}
With the above methods, we can remove the additive constant $c$ and achieve that: $\tilde g_{\hat \theta_{K,N}} \approx \log \frac{dP_1}{dP_0}$. 
Note that the constant, anyways, does not affect the \Ev-variable $E^{(K,\theta)}$ of our form \Cref{eq:e-var-ratio}, and only helps to interpret $g_{\hat \theta_{K,N}}$ as a log-likelihood ratio.
\end{Rem}

\begin{Rem}[Clipping - again]
\label{rem:clipping-emp}
 In order to prevent a learned \Ev-variable $E^{(K,{\hat \theta_{K,N}})}$, which might by chance become close to zero (log-unbounded from below), to quickly destroy the wealth process $W_t$, we could introduce a regularizing clipping parameter $\lambda \in [0,1]$, as described in \Cref{rem:clipping}:
\begin{align}
    E^{(K,\theta,\lambda)}(z^1,z^0_1,\dots,z^0_K) 
    &:= (1-\lambda) + \lambda \cdot E^{(K,\theta)}(z^1,z^0_1,\dots,z^0_K)\\
    & = (1-\lambda) + \lambda \cdot
    \frac{ \exp(g_\theta(z^1))}{\frac{1}{K+1}\lp \exp(g_\theta(z^1)) + \sum_{k=1}^K \exp(g_\theta(z^0_k)) \rp }.
\end{align}
This would, as mentioned in \Cref{rem:clipping}, restore the explicit type-II error bounds from \Cref{thm:type-II-error}. 
However, there are different ways to choose $\lambda$, as we will indicate in the following.

a.) A typical, simple and effective choice for $\lambda$ would be to just set $\lambda_1 := \frac{1}{2}$.
  
b.) Based on the observation, that we already have the deterministic upper bound:
 \begin{align}
     \log E^{(K,\theta)} & \le \log (K+1),
 \end{align}
 a symmetric lower bound could be achieved by the choice $\lambda_K := \frac{K}{K+1}$. Then we would just get:
\begin{align}
    E^{(K,\theta,\lambda_K)} 
    &= \frac{1}{K+1} + \frac{K}{K+1}\cdot E^{(K,\theta)} 
    = \frac{1}{K+1} +\frac{ \exp(g_\theta(z^1))}{\frac{1}{K}\lp \exp(g_\theta(z^1)) + \sum_{k=1}^K \exp(g_\theta(z^0_k)) \rp },
\end{align}
 with the symmetric lower and upper bounds:
 \begin{align}
    - \log (K+1) \le  \log E^{(K,\theta,\lambda_K)} & \le \log(K+1).
 \end{align}
 
 c.) Alternatively, we could jointly fit $\theta$ and $\lambda$, to maximize \Ev-power, similarly to before:
\begin{align}
   (\hat \theta_{K,N},\hat \lambda_{K,N}) :\in  \argmax_{\substack{\theta \in \Theta\\ \lambda \in [0,1]}}
    J_{K,N}(\theta,\lambda), &  &J_{K,N}(\theta,\lambda)&:= \frac{1}{N}\sum_{n=1}^N \log E^{(K,\theta,\lambda)}\lp z^1_n,(z^0_{n,k})_{k \in [K]} \rp.
\end{align}
Since $\lambda$ might, in this fitting procedure, still become too overconfident ($\hat \lambda_{K,N} \approx 1$), we could afterwards scale it down by a fixed hyperparameter $\rho \in [0,1)$, e.g.\ again $\rho_1=\frac{1}{2}$, $\rho_2=\frac{2}{3}$, $\rho_3=\frac{3}{4}$, $\rho_4=\frac{4}{5}$ or $\rho_K = \frac{K}{K+1}$, and put:
\begin{align}
    \tilde \lambda_{K,N} &:= \rho \cdot \hat \lambda_{K,N},
\end{align}
and use the \Ev-variable
\begin{align}
   E^{(K,\rho,N)} := E^{(K,\hat \theta_{K,N},\tilde \lambda_{K,N})},
\end{align}
which then always produces the deterministic bounds:
\begin{align}
    \log (1 - \rho) \le \log E^{(K,\rho,N)} \le \log (K+1).
\end{align}

d.) There are further options, like providing a whole schedule $(\lambda_t)_{t \in \N_0}$ over time, or even learning such a schedule (but only on past data). All these choices would not violate the conditional \Ev-variable conditions and all guarantees from \Cref{thm:type-I-error} and \Cref{thm:type-II-error} would still hold with the corresponding numbers.
\end{Rem}

\begin{Rem}[The choice of the number $K$]
There are two considerations for the choice of the number $K \ge 1$ that restrict the power of $E^{(K,\theta)}$.

a.) The first obstruction is the deterministic upper bound: \begin{align}
      E^{(K,\theta)} & \le K+1,
\end{align}
which restricts the \Ev-power to:
\begin{align}
  \E_{\Pbb_1}\lB \log E^{(K,\theta)}\rB & \le \log(K+1).
\end{align}
If we even want to have a chance to be approximately close to the upper limit of $\KL(P_1\|P_0)$ then at least the upper bound should be close to or even exceed it:
\begin{align}
    \log(K+1) & \gtrsim \KL(P_1\|P_0),
\end{align}
leading to the following condition for the choice of $K$:
\begin{align}
    K & \gtrsim \exp(\KL(P_1\|P_0)) -1 
    \approx \exp\lp \frac{1}{M} \sum_{m=1}^M \tilde g_{\hat \theta_{K,N}}(z^1_m)\rp - 1, & (z^1_m)_{m \in [M]} \overset{\text{i.i.d.}}{\sim} P_1.
\end{align}
which one could track and adjust during training.

b.) The second obstruction is the lower bound from \Cref{thm:approx-gro} for the optimized limit \Ev-variable:
\begin{align}
    \E_{\Pbb_1}[\log E^{(K,\hat \theta)}] \ge \quad \KL(P_1\|P_0) - \log\lp 1 + \frac{\chi^2(P_1\|P_0)}{K+1} \rp - \frac{1}{2}\epsilon^2.
\end{align}
If we wanted the term for the penalty for finite $K$ to become of size smaller than $\delta > 0$  we could require:
\begin{align}
  \delta & \overset{!}{\gtrsim}  \frac{\chi^2(P_1\|P_0)}{K} \ge \frac{\chi^2(P_1\|P_0)}{K+1} \ge \log\lp 1 + \frac{\chi^2(P_1\|P_0)}{K+1}\rp.
\end{align}
This would lead to the following criterion:
\begin{align}
    K   \gtrsim \frac{\chi^2(P_1\|P_0)}{\delta} 
    &\approx \frac{1}{\delta} \cdot \frac{1}{M}
    \sum_{m=1}^M \lp \exp(\tilde g_{\hat \theta_{K,N}}(z^0_m))-1\rp^2, & (z^0_m)_{m \in [M]} \overset{\text{i.i.d.}}{\sim} P_0 \\
    &\approx \frac{1}{\delta} \cdot \lp \frac{1}{M}
    \sum_{m=1}^M \exp(\tilde g_{\hat \theta_{K,N}}(z^1_m)) - 1\rp, & (z^1_m)_{m \in [M]} \overset{\text{i.i.d.}}{\sim} P_1.
\end{align}
As before, also this quantity could be checked and adjusted during training. 

Note that if we follow this advice and choose $\delta \le 1$, then we get with help of Jensen's inequality: 
\begin{align}
   \log(K+1) & \gtrsim \log\lp 1+ \frac{\chi^2(P_1\|P_0)}{\delta} \rp \ge \log\lp 1+ \chi^2(P_1\|P_0) \rp \ge \KL(P_1\|P_0),
\end{align}
which shows that we then would also satisfies point a.) at the same time. 

In summary, we would recommend to choose:
$K \propto \chi^2(P_1\|P_0)$.
\end{Rem}

\begin{Rem}[Alternative training objective]
Another popular and rather stable training objective in machine learning for learning density ratios comes from another variational form of the $\KL$-divergence, see \cite{Ngu10, Bel18, Poo019}: 
\begin{align}
 \KL(P_1\|P_0) &= \sup_{g} \lC \E_{P_1}[g] - \E_{P_0}[\exp(g)] + 1 \rC, \qquad \text{with unique maximizer:} \qquad g^*(z) = \log \frac{dP_1}{dP_0}(z).
\end{align}
Here, we then would, again, maximize the corresponding empirical version with samples $(z^1_n)_{n \in [N_1]} \overset{\text{i.i.d.}}{\sim} P_1$ and $(z^0_n)_{n \in [N_0]} \overset{\text{i.i.d.}}{\sim} P_0$:
\begin{align}
   \hat \theta_{N_1,N_0} &:\in \argmax_{\theta \in \Theta} I_{N_1,N_0}(\theta), 
   & I_{N_1,N_0}(\theta) &:= \frac{1}{N_1}\sum_{n=1}^{N_1} g_\theta(z^1_n) - \frac{1}{N_0}\sum_{n=1}^{N_0} \exp(g_\theta(z^0_n)) + 1.
\end{align}
The estimator is known to have less bias, but higher variance. 
After training, one can use $g_{\hat \theta_{N_1,N_0}}$ in our \Ev-variable from \Cref{eq:e-var-ratio} with some chosen $K \ge 1$.
\end{Rem}

The above concepts and constructions can also be combined with the following.

\begin{Rem}[Finite time horizon and inverse temperature]
Recall that in \Cref{sec:e-val-constr} we used the construction from \Cref{eq:E-val-K-h} to approximate the \Ev-variable from \Cref{eq:E-val-h-norm}:
\begin{align}
  \tilde h(z^1):= \frac{h(z^1)}{\E_{P_0}[h]} &\approx   \frac{h(z^1)}{\frac{1}{K+1}\lp h(z^1)+\sum_{k=1}^K h(z^0_k)\rp}  =: E^{(K,h)}(z^1,z^0_1,\dots, z^0_K),
\end{align}
where we then optimized $h$ in \Cref{thm:approx-gro} to receive: $h=\exp(g_{\hat \theta_{K,N}})$ with $g_{\hat \theta_{K,N}} \approx \log \frac{dP_1}{dP_0}$, which is approximately growth optimal in the long run and minimizes average rejection time, see \Cref{rem:rejection-time}.  
In contrast, following \cite{For26a}, we now want to consider the case where the time horizon $T \in \N$ for the real experiments $(x_t)_{t \in [T]} \overset{\text{i.i.d.}}{\sim} Q$ is a priori known and finite.\footnote{Also see \cite{Kon26} Cor.\ 4 for the interpretation of the tempered likelihood ratio as the optimal \Ev-variable w.r.t.\ the expected power utility function.} For this we want to fix $h$ from above, but we now modify it with an \emph{inverse temperature} regularization parameter $\beta \in \R_{>0}$:
\begin{align}
    h_\beta(z^1) := \frac{h(z^1)^\beta}{\E_{P_0}[h^\beta]} &\approx   \frac{h(z^1)^\beta}{\frac{1}{K+1}\lp h(z^1)^\beta+\sum_{k=1}^K h(z^0_k)^\beta\rp}  =: E^{(K,h,\beta)}(z^1,z^0_1,\dots, z^0_K),
\end{align}
in order to minimize the type-II error at the one future time $T$, when at present time $t \le T$, when we have already observed $W_{t-1}$:
\begin{align}
    \beta_{t-1} : \in \argmin_{\beta > 0} P_1\lp W_T \le \frac{1}{\alpha} \st \Fcal_{t-1}\rp. 
\end{align}
Assume for now, that we are in the idealized case, where we have directly used $h_\beta$ as \Ev-variables; then the above reduces to:
\begin{align}
P_1\lp W_T \le \frac{1}{\alpha} \st \Fcal_{t-1}\rp 
&= P_1\lp \sum_{n=t}^T \log h_\beta(z_n^1) \le \log\frac{1}{\alpha} - \log W_{t-1} \st \Fcal_{t-1}\rp \\
&= P_1\Bigg( \frac{1}{T-t+1}\sum_{n=t}^T \log h(z_n^1) \le \frac{1}{\beta}\bigg( \frac{\log\frac{1}{\alpha} - \log W_{t-1}}{T-t+1} + \log \E_{P_0}[h^\beta] \bigg)  \,\Bigg|\, \Fcal_{t-1} \Bigg).
\end{align}
Note that only the rhs depends on $\beta$, so minimizing that type-II error reduces to minimizing the rhs w.r.t.
 $\beta$, which we only need for the case when we have not rejected $H_1$ yet, that is, if $W_{t-1} \le \frac{1}{\alpha}$. We  thus arrive at the exact objective:
\begin{align}
    \beta_{t-1} \in \argmin_{\beta > 0} \frac{1}{\beta} \Bigg(\underbrace{\frac{(\log \frac{1}{\alpha} - \log W_{t-1})_+ }{T-t+1}}_{=:\ell_{t-1}} + \underbrace{\log \E_{P_0}\lB h^\beta\rB}_{=:\psi(\beta)} \Bigg),
\end{align}
where we interpret $\ell_{t-1}$ as the residual target evidence for rejecting $H_1$ per remaining time step until time $T$. 
The optimization problem can then numerically be solved with Newton iteration:
\begin{align}
    \beta^{(i+1)} := \beta^{(i)} - \frac{\beta^{(i)} \cdot \psi'(\beta^{(i)}) - \psi(\beta^{(i)}) - \ell_{t-1}}{\beta^{(i)} \cdot \psi''(\beta^{(i)})}, \qquad \text{ with: }
\end{align}
\begin{align}
        \psi(\beta) &= \log \E_{P_0}[h^\beta], & 
        \psi'(\beta) &= \frac{\E_{P_0}[h^\beta\log h]}{\E_{P_0}[h^\beta]}  & 
        \psi''(\beta) &= \frac{\E_{P_0}[h^\beta(\log h)^2]}{\E_{P_0}[h^\beta]} - \psi'(\beta)^2,
\end{align}
where each expectation value is replaced by the empirical counterparts based on the same one batch of i.i.d.\ samples $(z^0_m)_{m \in [M]} \overset{\text{i.i.d.}}{\sim} P_0$ and with $h=\exp(g_{\hat \theta_{K,N}})$.
After convergence, $i \to \infty$, practically after $3-5$ rounds of iterations, one has found an approximate $\beta_{t-1}:=\beta^{(\infty)}$, which one can use, at time $t \in [T]$, for the conditional \Ev-variable: 
\begin{align}
  E_t(z^1,z^0_1,\dots,z^0_K) 
&:= \frac{ \exp(\beta_{t-1}\cdot g_{\hat \theta_{K,N}}(z^1))}{\frac{1}{K+1}\lp \exp(\beta_{t-1}\cdot g_{\hat \theta_{K,N}}(z^1)) + \sum_{k=1}^K \exp(\beta_{t-1}\cdot g_{\hat \theta_{K,N}}(z^0_k)) \rp }.
\end{align}  
At the next time step $t+1$, this can be repeated, leading to an adaptive predictive sequence $(\beta_t)_{t \in [T]}$ of inverse temperatures. 
This, of course, can be combined with the clipping regularization from \Cref{rem:clipping-emp}. Note that, even with these $\beta$ values in place, each of which is only dependent on their own past, the anytime type-I error guarantees from \Cref{thm:type-I-error} still hold.
\end{Rem}

\section{An alternative testing scenario}

Consider the scenario, where we want to test:
\begin{align}
    H_0: Q= P_0, \qquad \text{ vs. } \qquad  H_1: Q \neq P_0,
\end{align}
For this we pre-train $E^{(K,\theta)}$ as before with help of some alternative $P_1 \neq P_0$ first, if possible. However, to adjust to the real data distribution coming from $Q$, which we suspect to deviate from $P_0$, we will then train $E^{(K,\theta)}$ on all \emph{past} observations $x_n \sim Q$ with $n =1, \dots, t-1$, explicitly excluding $x_t$ (and futute $x_n$) for training at time $t$, and sufficiently many simulations from $P_0$ as before (and/or stored simulations from the past). 
More concretely, at each time $t \ge 1$, we optimize the empirical objective with data from times until $t-1$:
\begin{align}
 (\hat \theta_{K,t-1}, \hat \lambda_{K,t-1}) &:\in   \argmax_{\substack{\theta \in \Theta\\ \lambda \in [0,1]}} \lC \frac{1}{t-1}\sum_{n=1}^{t-1} \log E^{(K,\theta,\lambda)}\lp x_n,(z^0_{n,k})_{k \in [K]} \rp - \kappa \cdot R_{K,t-1}(\theta) \rC,
\end{align}
and plug the new point $x_t \sim Q$ together with fresh simulations $(z^0_{t,k})_{k \in [K]} \overset{\text{i.i.d.}}{\sim} P_0$ into it:
\begin{align}
    E_t &:= E^{(K,\hat \theta_{K,t-1}, \tilde \lambda_{K,t-1},\beta_{t-1})}\lp x_t,(z^0_{t,k})_{k \in [K]} \rp.
\end{align}
This will still be a valid conditional \Ev-variable for $H_0$ and the anytime type-I error guarantees will still hold. 

Note that the trained \Ev-variable $E^{(K,\theta)}$ will then rather learn to approximate $\frac{dQ}{dP_0}$ than $\frac{dP_1}{dP_0}$, which, in this testing scenario, is more optimal, if $H_1$ is true.

\section{Discussion}

For the simulation-based hypothesis testing setting, $H_0: Q = P_0$ vs.\ $H_1: Q = P_1$, where we do not have densities and can only sample from $P_0$ and $P_1$, we constructed a sequential test based on an \Ev-test martingale, which comes with anytime-valid type-I error guarantees, approximate growth optimality, geometrically decaying type-II error bounds, and asymptotic power one.

Future work could address further generalizations for the \emph{anytime-valid simulation-based hypothesis testing} setting, where we have, instead of simple,  \emph{composite hypothesis classes} $\Hcal_0, \Hcal_1 \ins \Pcal(\Omega)$, leading to the statistical test $H_0: Q \in  \Hcal_0$ vs.\ $H_1: Q \in \Hcal_1$, and/or the setting of (filtration-adapted) \emph{dependent data streams} $(X_t)_{t \in \N_1} \sim Q$, instead of i.i.d.\ data. 
Furthermore, a more realistic \emph{approximate growth optimality theorem} could be proven for more realistic norms, non-compact $\Omega$ and $\Theta$, and (universally approximating) function classes of more practical relevance.

\newcommand{\etalchar}[1]{$^{#1}$}

\newpage
\appendix
\section*{Appendix}
\addcontentsline{toc}{section}{Appendix}

\section{Theorems}

\begin{Thm}[Ville's inequality, see \cite{Vil39}]
    \label{thm:ville}
    Let $(W_t)_{t \in \N_0}$ be a non-negative supermartingale, $W_t:\, (\Omega,\Bcal,P_0) \to [0,\infty]$, $t \in \N_0$, adapted to the filtration $\Fcal=(\Fcal_t)_{t \in \N_0}$, $\Fcal_t \ins \Bcal$ and w.r.t.\ probability measure $P_0$. 
    Then for every $\alpha \in (0,1]$ we have the inequality:
    \begin{align}
        P_0\lp \exists t \in \N_0.\; W_t > \frac{1}{\alpha} \rp \le \alpha \cdot \E_{P_0}[W_0].
    \end{align}
\end{Thm}

\begin{Thm}[Azuma–Hoeffding inequality, see \cite{Azu67}]
\label{thm:azuma}
Let $(L_t)_{t \in \N_0}$ be sequence of random variables $L_t:\, (\Omega,\Bcal,P_1) \to \R$, $t \in \N_0$, $L_0:=0$, adapted to the filtration $\Fcal=(\Fcal_t)_{t \in \N_0}$,  $\Fcal_t \ins \Bcal$ and w.r.t.\ probability measure $P_1$. Further assume that for every $t \in \N_1$ there are constants $a_t, b_t, c_t \in \R$ such that:
\begin{align}
    a_t &\le L_t \le b_t \quad P_1\text{-a.s.}, & 
    c_t & \le \E_{P_1}[L_t|\Fcal_{t-1}]
    \quad P_1\text{-a.s.}.
\end{align}
Then for every $\ell \in \R$ and $t \in \N_1$ we have:
\begin{align}
    P_1\lp \sum_{n=1}^t L_n \le \ell \rp & 
    \le \exp\lp - \frac{2\lp \sum_{n=1}^t c_n - \ell\rp_+^2}{\sum_{n=1}^t (b_n-a_n)^2} \rp.
\end{align}
If, in particular, $b_t-a_t \le d >0$ and $c_t \ge c$ for all $t \in \N_1$ then we get:
\begin{align}
    P_1\lp \sum_{n=1}^t L_n \le \ell \rp & 
    \le \exp\lp - 2t \cdot \frac{\lp c - \frac{1}{t}\ell \rp_+^2}{d^2} \rp.
\end{align}  
\end{Thm}

\section{Proofs}
\label{app:proof}

\begin{Prf}[\Cref{prp:optimal-e-power-k}]
    \label{prf:prp:optimal-e-power-k}
 In the following we abbreviate $\Xcal:=\Omega^{K+1}$ and 
$x=(z^1,z^0_1,\dots,z^0_K) \in \Xcal$.
We abbreviate: $\Ycal := \lC 0,1,\dots,K\rC$ and the uniform distribution on $\Ycal$:
\begin{align}
    P(Y = k) &:= \frac{1}{K+1}.
\end{align}
For $k \in \Ycal$ we then abbreviate the probability distribution on $\Xcal$ with random variable $X=(X_0,\dots,X_K)$:
\begin{align}
    \Pbb(X|Y=k) &:= P_0(X_0) \otimes \cdots \otimes P_0(X_{k-1}) \otimes P_1(X_k) \otimes P_0(X_{k+1}) \otimes \cdot \otimes P_0(X_K).
\end{align}
We further put:
\begin{align}
    Q^{(K,h)}(Y=k|X=x) &:= \frac{h(x_k)}{ \sum_{j=0}^K h(x_j)}.
\end{align}
Then we have:
\begin{align}
    \E_{P(X|Y) \otimes P(Y)}\lB \log Q^{(K,h)}(Y|X) \rB
    &= \frac{1}{K+1} \sum_{k=0}^K \E_{P(X|Y=k)} \lB \log \lp \frac{h(X_k)}{ \sum_{j=0}^K h(X_j)} \rp \rB \\
    &= \E_{P(X|Y=0)} \lB \log \lp \frac{h(X_0)}{ \sum_{j=0}^K h(X_j)} \rp \rB \\
    &= \E_{\Pbb_1}\lB \log E^{(K,h)}\rB - \log(K+1).
\end{align}
With this we get:
\begin{align}
    \argmax_{h} \E_{\Pbb_1}\lB \log E^{(K,h)}\rB 
    &= \argmax_{h} \E_{P(X|Y) \otimes P(Y)}\lB \log Q^{(K,h)}(Y|X) \rB \\
    &= \argmin_{h} \E_{P(X)} \lB \KL(P(Y|X)\|Q^{(K,h)}(Y|X)) \rB.
\end{align}
Since $\KL$ is a proper divergence, the rhs is minimized if and only if it equals zero if and only if:
\begin{align}
    P(Y|X) &= Q^{(K,h)}(Y|X) \qquad P(X)\text{-a.s.}
\end{align}
Since we assumed $P_1 \ll P_0$, we also have $P(X|Y=k) \ll P_0^{\otimes(K+1)}$ and $P(X) \ll P_0^{\otimes(K+1)}$. So we can express the lhs via Bayes's theorem in terms of densities $p(x|k)$ and $p(x)$:
\begin{align}
    P(Y=k|X=x) &= \frac{p(x|k)}{p(x)} P(Y=k)
    = \frac{\frac{dP_1}{dP_0}(x_k)}{\frac{1}{K+1}\sum_{j=0}^K \frac{dP_1}{dP_0}(x_j)} \frac{1}{K+1} = \frac{\frac{dP_1}{dP_0}(x_k)}{\sum_{j=0}^K \frac{dP_1}{dP_0}(x_j)}.
\end{align}
So, the condition for minimization holds if and only if for $P(X)$-almost-all $x \in \Xcal$ and all $k \in \Ycal$ we have:
\begin{align}
\frac{\frac{dP_1}{dP_0}(x_k)}{\sum_{j=0}^K \frac{dP_1}{dP_0}(x_j)} 
= \frac{h(x_k)}{ \sum_{j=0}^K h(x_j)}. \label{eq:ratio-normed}
\end{align}
Taking ratios for $k,l \in \Ycal$ gives $P(X)$-almost-all $x \in \Xcal$:
\begin{align}
h(x_k) \cdot \frac{dP_1}{dP_0}(x_l) &= h(x_l) \cdot \frac{dP_1}{dP_0}(x_k).
\end{align}
Inegrating over $P_0(dx_l)$ gives for every $k \in \Ycal$:
\begin{align}
    h(x_k) &=  h(x_k) \cdot \underbrace{\E_{P_0}\lB \frac{dP_1}{dP_0}\rB}_{=1} = 
    \underbrace{\E_{P_0}\lB h \rB}_{=:c >0} \cdot \frac{dP_1}{dP_0}(x_k) =  c \cdot \frac{dP_1}{dP_0}(x_k), 
\end{align}
which implies that for some $c >0 $, $P_0$-almost-all $z \in \Omega$:
\begin{align}
    h(z) &=    c \cdot \frac{dP_1}{dP_0}(z). \label{eq:ratio-const}
\end{align}
Note that \Cref{eq:ratio-const} also directly implies \Cref{eq:ratio-normed}. 
This then shows that such an $h$ maximizes the \Ev-power $\E_{\Pbb_1}\lB \log E^{(K,h)}\rB$.
This shows the equivalence.
\end{Prf}

\begin{Prf}[\Cref{thm:approx-gro}]
\label{prf:thm:approx-gro}
1.) In the following we abbreviate $\Xcal:=\Omega^{K+1}$ and 
$x=(z^1,z^0_1,\dots,z^0_K) \in \Xcal$.
First note that we have for every $x \in \Xcal$ the uniform bound:
\begin{align}
     E^{(K,\theta)}(x) &=  \frac{ \exp(g_\theta(z^1))}{\frac{1}{K+1}\lp \exp(g_\theta(z^1)) + \sum_{k=1}^K \exp(g_\theta(z^0_k)) \rp } \le K+1.
\end{align}
We further abbreviate:
\begin{align}
    \ell_\theta(x) &:= \log E^{(K,\theta)}(x) = g_\theta(z^1) - \log \lp \exp(g_\theta(z^1)) + \sum_{k=1}^K \exp(g_\theta(z^0_k)) \rp + \log (K+1).
\end{align}
Note that we have the bounds:
\begin{align}
    \log(K+1) \ge \ell_\theta(x) & \ge g_\theta(z^1) - \max\lp g_\theta(z^1), g_\theta(z^0_1), \dots, g_\theta(z^0_K) \rp \\
    &\ge - g_{\theta,-}(z^1) - \max_{k \in [K]} g_{\theta,+}(z^0_k),
\end{align}
which implies the two bounds:
\begin{align}
    \ell_{\theta,+}(x) & \le \log(K+1) < \infty, &
    \ell_{\theta,-}(x) & \le g_{\theta,-}(z^1) + \max_{k \in [K]} g_{\theta,+}(z^0_k).
\end{align}
For the latter we take $\sup_{\theta \in \Theta}$ and then $\E_{\Pbb_1}$ and get:
\begin{align}
   \E_{\Pbb_1}\lB\sup_{\theta \in \Theta}\ell_{\theta,-}(X)\rB 
   & \le \E_{P_1}\lB \sup_{\theta \in \Theta} g_{\theta,-}(Z^1)\rB + \E_{P_0}\lB\max_{k \in [K]} \sup_{\theta \in \Theta} g_{\theta,+}(Z^0_k)\rB \\
   & \le \E_{P_1}\lB \sup_{\theta \in \Theta} g_{\theta,-}(Z^1)\rB + \sum_{k \in [K]} \E_{P_0}\lB \sup_{\theta \in \Theta} g_{\theta,+}(Z^0_k)\rB \\
   & \overset{\ref{it:unif-int-envs}}{<} \infty,
\end{align}
where the latter holds by assumption \ref{it:unif-int-envs} about the integrable envelopes. Together we get the finite integrable envelope bound:
\begin{align}
    \E_{\Pbb_1}\lB \sup_{\theta \in \Theta}|\ell_{\theta}(X)| \rB & < \infty. \label{eq:log-e-env-integrable}
\end{align}

2.) Also note that by assumption \ref{it:cont}, $\ell_\theta(x)$ is continuous in $\theta$ for every fixed $x \in \Xcal$. By the integrability condition from \Cref{eq:log-e-env-integrable} we can use the dominated convergence theorem to show that also:
\begin{align}
    J_K(\theta) &= \E_{\Pbb_1}[\ell_\theta(X)]
\end{align}
is continuous in $\theta \in \Theta$. Since $\Omega$ is compact by assumption \ref{it:compact} we even get that $J_K$ is uniformly continuous.

3.) We now want to prove the \emph{uniform law of large numbers}:
\begin{align}
    \sup_{\theta \in \Theta} \lI J_{K,N}(\theta) - J_K(\theta) \rI \overset{N \to \infty}{\longrightarrow} 0 \qquad \Pbb_1\text{-a.s.}
\end{align}
To show this, let $\epsilon > 0$ be fixed.

a.) Now recall:
\begin{align}
    J_{K,N}(\theta) & = \frac{1}{N}\sum_{n=1}^N \ell_\theta(X_n), & J_K(\theta) &= \E_{\Pbb_1}\lB J_{K,N}(\theta)\rB.
\end{align}
Since $J_K$ is uniformly continuous, there exists $\eta_1 > 0$ such that for all $\theta',\theta" \in \Theta$ we have:
\begin{align}
    d(\theta',\theta") \le \eta_1 & \implies |J_K(\theta') - J_K(\theta") | < \epsilon/4.
\end{align}

b.) As a next step, for $\eta > 0$ and $x \in \Xcal$ define:
\begin{align}
    D_\eta(x) &:= \sup_{\substack{\theta',\theta" \in \Theta\\d(\theta',\theta")\le \eta}} |\ell_{\theta'}(x) - \ell_{\theta"}(x) |.
\end{align}
Note that $D_\eta(x)$ in monotone non-decreasing in $\eta$ for each fixed $x \in \Xcal$ with:
\begin{align}
    \lim_{\eta \to 0}  D_\eta(x)  & = 0,
\end{align}
due to the uniform continuity of $\ell_{\theta}(x)$ in $\theta$, combining the assumptions of continuity \ref{it:cont} and compactness  \ref{it:compact}. 
Further note that we have the bound:
\begin{align}
  0 \le  D_\eta(x)  & \le 2 \cdot \sup_{\theta \in \Theta}|\ell_{\theta}(x)|,
\end{align}
where the upper bound is integrable by \Cref{eq:log-e-env-integrable}.
The dominated convergence theorem thus implies:
\begin{align}
    \lim_{\eta \to 0} \E_{\Pbb_1}\lB D_\eta(X) \rB & = 0.
\end{align}
So, we can choose $\eta >0$ small enough and with $ \eta_1 > \eta > 0$ such that:
\begin{align}
    \E_{\Pbb_1}\lB D_\eta(X) \rB & < \epsilon/4.
\end{align}
By the strong law of large numbers we have the $\Pbb_1$-almost-sure convergence:
\begin{align}
    \frac{1}{N}\sum_{n=1}^N D_\eta(X_n) &\overset{N\to\infty}{\longrightarrow} \E_{\Pbb_1}\lB D_\eta(X) \rB  < \epsilon/4.
\end{align}
So, for $\Pbb_1$-almost-all realizations $\omega \in \Omega$ there exists an $N_0(\omega) \in \N$ such that for all $N \ge N_0(\omega)$ we have:
\begin{align}
    \frac{1}{N}\sum_{n=1}^N D_\eta(X_n(\omega)) &  < \epsilon/2.
\end{align}

c.) By assumption \ref{it:compact} the space $\Omega$ is a compact metric space with some metric $d$. So, there are finitely many $\theta_1,\dots, \theta_M \in \Omega$ with:
\begin{align}
    \Omega &\ins \bigcup_{m \in [M]} B(\theta_m,\eta),
\end{align}
where $\eta>0$ was fixed in the previous points and $B(\theta_m,\eta)$ is the (closed) ball of radius $\eta$ around $\theta_m$. So, for every $\theta \in \Omega$ we can pick an index $j(\theta) \in [M]$ such that:
\begin{align}
    d(\theta,\theta_{j(\theta)}) \le \eta,
\end{align}
 which then implies:
\begin{align}
    |J_K(\theta) - J_K(\theta_{j(\theta)}) | < \epsilon/4.
\end{align}

d.) Now consider $\theta \in \Omega$ and $m=j(\theta) \in [M]$. Then we have:
\begin{align}
    \lI J_{K,N}(\theta) - J_{K,N}(\theta_m) \rI
    & = \lI \frac{1}{N} \sum_{n=1}^N \lp \ell_\theta(X_n) - \ell_{\theta_m}(X_n) \rp \rI \\
    & \le  \frac{1}{N} \sum_{n=1}^N \lI \ell_\theta(X_n) - \ell_{\theta_m}(X_n) \rI \\
    & \le \frac{1}{N} \sum_{n=1}^N D_\eta(X_n)  \\
    & < \epsilon/2,
\end{align}
for $\Pbb_1$-almost-all realizations $\omega \in \Omega$ and all $N \ge N_0(\omega)$.

e.) The strong law of large numbers shows that for every $m \in [M]$ we have the $\Pbb_1$-almost-sure convergence:
\begin{align}
    J_{K,N}(\theta_m) & \overset{N\to\infty}{\longrightarrow} J_K(\theta_m),
\end{align}
and since $M$ is finite also the $\Pbb_1$-almost-sure convergence:
\begin{align}
   \max_{m \in [M]} \lI J_{K,N}(\theta_m) - J_K(\theta_m) \rI & \overset{N\to\infty}{\longrightarrow} 0,
\end{align}
So, for $\Pbb_1$-almost-all realizations $\omega \in \Omega$ and all $N \ge N_0(\omega)$ we have:
\begin{align}
   \max_{m \in [M]} \lI J_{K,N}(\theta_m) - J_K(\theta_m) \rI & < \epsilon/4.
\end{align}

f.) Putting everything together shows that for fixed $\epsilon >0$ and $\Pbb_1$-almost-all realizations $\omega \in \Omega$, all $\theta \in \Theta$ and all $N \ge N_0(\omega)$, and with $m=j(\theta) \in [M]$, we have:
\begin{align}
    \lI J_{K,N}(\theta) - J_K(\theta) \rI 
    & \le \underbrace{\lI J_{K,N}(\theta) - J_{K,N}(\theta_m) \rI}_{\le \frac{1}{N} \sum_{n=1}^N D_\eta(X_n) < \epsilon/2}
    + \underbrace{\lI J_{K,N}(\theta_m) - J_K(\theta_m) \rI}_{< \epsilon/4}
    + \underbrace{\lI J_K(\theta_m) - J_K(\theta) \rI}_{<\epsilon/4} 
     < \epsilon.
\end{align}
Note that, due to the uniform bound involving $D_\eta$, the bound $N_0(\omega)$ depends only on the realization $\omega \in \Omega$ (and $\epsilon$), but crucially not on $\theta$.
We thus get:
\begin{align}
    \sup_{\theta \in \Theta} \lI J_{K,N}(\theta) - J_K(\theta) \rI 
    & \le  \epsilon.
\end{align}
This thus shows the \emph{uniform law of large numbers}:
\begin{align}
   \Delta_{K,N} := \sup_{\theta \in \Theta} \lI J_{K,N}(\theta) - J_K(\theta) \rI \overset{N \to \infty}{\longrightarrow} 0 \qquad \Pbb_1\text{-a.s.}
\end{align}

4.) We now further have the (near-)maximizing sequence $(\hat \theta_{K,N})_{N \in \N}$, i.e.\ for all $N \in \N$:
\begin{align}
    J_{K,N}(\hat \theta_{K,N}) & \ge \sup_{\theta \in \Theta}  J_{K,N}(\theta) - \xi_{K,N} &  \limsup_{N\to \infty}\xi_{K,N} = \xi_K \in \R_{\ge 0} \qquad \Pbb_1\text{-a.s.}
\end{align}
By definition of $\Delta_{K,N}$ we now have:
\begin{align}
 \sup_{\theta \in \Theta}  J_K(\theta) \ge  J_K(\hat \theta_{K,N}) & = J_{K,N}(\hat \theta_{K,N}) + J_K(\hat \theta_{K,N}) - J_{K,N}(\hat \theta_{K,N}) \\
    & \ge J_{K,N}(\hat \theta_{K,N}) - \lI J_K(\hat \theta_{K,N}) - J_{K,N}(\hat \theta_{K,N}) \rI   \\
    &\ge J_{K,N}(\hat \theta_{K,N}) - \Delta_{K,N} \\
    &\ge \sup_{\theta \in \Theta}  J_{K,N}(\theta) - \Delta_{K,N} - \xi_{K,N} \\
    & \overset{(*)}{\ge} \sup_{\theta \in \Theta}  J_K(\theta) - 2\Delta_{K,N} - \xi_{K,N}.
\end{align}
To get $(*)$ observe that we have:
\begin{align}
    J_{K,N}(\theta) & = J_K(\theta) + J_{K,N}(\theta) - J_K(\theta)  \\
    & \ge J_K(\theta) - \lI J_{K,N}(\theta) - J_K(\theta) \rI   \\
    &\ge J_K(\theta) - \Delta_{K,N},
\end{align}
which gives:
\begin{align}
    \sup_{\theta \in \Theta} J_{K,N}(\theta) & \ge \sup_{\theta \in \Theta} J_K(\theta) - \Delta_{K,N}.
\end{align}
Altogether, we have:
\begin{align}
 \sup_{\theta \in \Theta}  J_K(\theta) \ge  J_K(\hat \theta_{K,N}) & \ge \sup_{\theta \in \Theta}  J_K(\theta) - 2\Delta_{K,N} - \xi_{K,N} \qquad \Pbb_1\text{-a.s.},
\end{align}
with:
\begin{align}
 \limsup_{N \to \infty}\Delta_{K,N} & =0, & \limsup_{N \to \infty} \xi_{K,N} &= \xi_K, \qquad \Pbb_1\text{-a.s.}.
\end{align}
This then implies:
\begin{align}
   \liminf_{N \to \infty} J_K(\hat \theta_{K,N}) &\ge \sup_{\theta \in \Theta}  J_K(\theta) - \xi_K \qquad \Pbb_1\text{-a.s.}
\end{align}
This shows another main claim of \Cref{thm:approx-gro}.

5.) g.) For the final main claim, we need to show the inequality:
\begin{align}
   J_K(g) & \ge J_K(g_c^*) - \frac{1}{2}\epsilon^2, \qquad \text{ whenever }\qquad \|g-g^*_c\|_\infty \le \epsilon.
\end{align}
For this we introduce the curve in function space, $s \in \R$:
\begin{align}
    g_s(z) &:= g^*_c(z) + s \cdot v(z), & v(z) &:=  g(z) - g^*_c(z) \in [-\epsilon,\epsilon].
\end{align}
and 
\begin{align}
     x&:=(z^1,z^0_1,\dots,z^0_K), &
    \pi_k(s,x) &:= \frac{\exp(g_s(x_k))}{\sum_{j=0}^K \exp(g_s(x_j))}, .
\end{align}
We then introduce:
\begin{align}
  J(s) := J_K(g_s) 
    & = \E_{\Pbb_1}\lB g_s(X_0) - \log\lp \sum_{k=0}^K \exp(g_s(X_k)) \rp  \rB + \log(K+1).
\end{align}
With this we get the derivatives:
\begin{align}
    J'(s) &= \E_{\Pbb_1}\lB v(X_0) - \sum_{k=0}^K \pi_k(s,X) \cdot  v(X_k) \rB, \\
    J''(s) &= -\E_{\Pbb_1}\lB \sum_{k=0}^K \pi_k(s,X) \cdot \lp v(X_k) - \sum_{j=0}^K \pi_j(s,X) \cdot  v(X_j) \rp \cdot  v(X_k) \rB \\
    &= -\E_{\Pbb_1}\lB \sum_{k=0}^K \pi_k(s,X) \cdot v(X_k)^2 - \lp \sum_{k=0}^K \pi_k(s,X) \cdot  v(X_k)  \rp^2 \rB \\
    &= -\E_{\Pbb_1}\lB \sum_{k=0}^K \pi_k(s,X) \cdot v(X_k)^2 \rB + \E_{\Pbb_1}\lB\lp \sum_{k=0}^K \pi_k(s,X) \cdot  v(X_k)  \rp^2 \rB \\
    &\ge -\E_{\Pbb_1}\lB \sum_{k=0}^K \pi_k(s,X) \cdot \underbrace{v(X_k)^2}_{\le \epsilon^2} \rB \\
    &\ge - \epsilon^2.
\end{align}
Then note that $g_0=g^*_c$ is a maximizer of $J_K$, so at $s=0$ we get vanishing first derivative:
\begin{align}
    J'(0) & = 0.
\end{align}
Then we consider the Taylor expansion around $s=0$ and get:
\begin{align}
  J_K(g) =  J(1) &= J(0) + \underbrace{J'(0)}_{=0} + \int_0^1 (1-u)\cdot \underbrace{J''(u)}_{\ge - \epsilon^2}\,du \\ & \ge J(0) - \epsilon^2 \cdot \int_0^1 (1-u)\,du \\
    & = J(0) - \frac{1}{2}\epsilon^2 \\
    & = J_K(g_c^*) - \frac{1}{2}\epsilon^2,
\end{align}
implying the inequality:
\begin{align}
   J_K(g) & \ge J_K(g_c^*) - \frac{1}{2}\epsilon^2, \qquad \text{ whenever }\qquad \|g-g^*_c\|_\infty \le \epsilon.
\end{align}
By the approximation assumption \ref{it:approx-class} we thus get the inequalities :
\begin{align}
 \KL(P_1\|P_0) \ge J_K(g_c^*) \ge   \sup_{\theta \in \Theta} J_K(g_\theta)  & \ge J_K(g_c^*) - \frac{1}{2}\epsilon^2.
\end{align}

h.) Finally, if we assume $\chi^2(P_1\|P_0) < \infty$, then a simple Jensen inequality shows:
\begin{align}
    J_K(g_c^*) &= \E_{\Pbb_1}\lB \log \lp 
    \frac{ \frac{dP_1}{dP_0}(Z^1)}{\frac{1}{K+1}\lp \frac{dP_1}{dP_0}(Z^1) + \sum_{k=1}^K \frac{dP_1}{dP_0}(Z^0_k)\rp} \rp\rB \\
    &= \KL(P_1\|P_0) - \E_{\Pbb_1}\lB \log \lp 
    \frac{1}{K+1}\lp \frac{dP_1}{dP_0}(Z^1) + \sum_{k=1}^K \frac{dP_1}{dP_0}(Z^0_k)\rp\rp\rB \\
    &\ge \KL(P_1\|P_0) - \log \E_{\Pbb_1}\lB  
    \frac{1}{K+1}\lp \frac{dP_1}{dP_0}(Z^1) + \sum_{k=1}^K \frac{dP_1}{dP_0}(Z^0_k)\rp\rB \\
    &= \KL(P_1\|P_0) - \log \lp  
    \frac{1}{K+1}\lp \E_{P_1}\lB \frac{dP_1}{dP_0}(Z^1) \rB + \sum_{k=1}^K \E_{P_0}\lB \frac{dP_1}{dP_0}(Z^0_k) \rB\rp \rp \\
    &= \KL(P_1\|P_0) - \log \lp  
    \frac{1}{K+1}\lp 1 +\chi^2(P_1\|P_0) + \sum_{k=1}^K 1\rp \rp \\
    &= \KL(P_1\|P_0) - \log \lp  
    1 +\frac{\chi^2(P_1\|P_0)}{K+1} \rp.
\end{align}
With this all claims are shown in \Cref{thm:approx-gro}.
\end{Prf}

\end{document}